\documentclass[conference]{IEEEtran}
\usepackage{algorithmic}
\usepackage{array}
 \usepackage{verbatim}
\usepackage{amsthm}
\usepackage{amsmath,amssymb}
\usepackage{graphicx}
\usepackage{listings}
\usepackage{tikz}

\usepackage{color}
\usepackage{url}

\definecolor{olivegreen}{rgb}{0.2,0.8,0.5}
\definecolor{grey}{rgb}{0.5,0.5,0.5}

\lstdefinelanguage{ttl}{
sensitive=true,
showspaces=false,
showstringspaces=false,
keywords={author, birthPlace, birthDate, broader,subject, label, completionDate, type, genre, philosophicalSchool,publicationDate, mainInterest, movement},
morecomment=[l][\color{grey}]{--},
morestring=[b][\color{blue}]\",
morecomment=[s][\color{olivegreen}]{<}{>},
alsoletter={-,<,>},
emph={dbo,dbr,dc, Category, rdf, rdfs, skos}, emphstyle=\itshape,
emph={[2]fun,cat, lincat,lin},emphstyle={[2]\color{red}}
}

\ifCLASSOPTIONcompsoc
  \usepackage[caption=false,font=normalsize,labelfont=sf,textfont=sf]{subfig}
\else
  \usepackage[caption=false,font=footnotesize]{subfig}
\fi

\begin{document}
\title{Assisting Drivers During Overtaking Using Car-2-Car Communication and Multi-Agent Systems}

\author{\IEEEauthorblockN{Adrian Groza\IEEEauthorrefmark{1},
Calin Cara\IEEEauthorrefmark{1},
Sergiu Zaporojan\IEEEauthorrefmark{3}, 
Igor Calmicov\IEEEauthorrefmark{3}
\IEEEauthorblockA{\IEEEauthorrefmark{1}Department of Computer Science, Technical University of Cluj-Napoca, Romania\\
Adrian.Groza@cs.utcluj.ro, Calin.Cara@cs-gw.utcluj.ro\\ 
\IEEEauthorrefmark{3}Department of Computer Science, Technical University of Moldova, Moldova\\
Sergiu.Zaporojan@cs.utm.md,Igor.Calmicov@cs.utm.md\\ 
}
}
}

\maketitle

\begin{abstract}
A warning system for assisting drivers during overtaking maneuvers is proposed. 
The system relies on Car-2-Car communication technologies and multi-agent systems. 
A protocol for safety overtaking is proposed based on ACL communicative acts.
The mathematical model for safety overtaking used Kalman filter to minimize localization error.
\end{abstract}



\section{Introduction}
Smart vehicles currently share with Internet of Things the first rank regarding expectations, as demonstrated by the Gartner’s Hype (http://www.gartner.com/newsroom/id/3114217). 
Smart vehicles represent the major change in the 2015 Hype Cycle for Emerging Technologies, as this technology has shifted from pre-peak to peak of the Hype Cycle. The main rationale is that all major automotive companies are putting smart vehicles on their near-term road-maps.
At the EU level, European Commission (EC) is getting involved and promoting several Field Operational Test, while European Telecommunications Standards Institute (ETSI) has finally standardized higher layer networking protocols.

The main bottleneck is that C2X technology has just solved low level aspects with respect to ad-hoc networks or regulatory norms and standards, with much work remained at the application layer. 
In this paper, advances in multi-agent systems (communication protocols, speech acts) are exploited to deploy cooperative, communication based active safety application.

The remaining of this paper is structured as follows:
Section~\ref{sec:model} formalises the mathematical model for vehicle overtaking.
Section~\ref{sec:agents} details the communication protocol and the experiments run based on multi-agent systems.
Discussion appears section in\ref{sec:discussion}, while section~\ref{sec:conclusion} concludes the paper. 

\section{Cooperative safety overtaking}
\label{sec:model}
This section formalises the mathematical model for overtaking and the approach for minimizing localisation error of the vehicles envolved in overtaking.

\subsection{$2^+$ Overtaking Model}
In the $2^+$ overtaking, multiple vehicles which have to be overtaken in a single maneuver. 
The overtaking car will have to travel a longer distance on the opposite road before returning to the original lane.
A novel model for the $\mathit{2^+}$ overtaking maneuver is introduced in this section. 
The model is an extension of the mathematical model described in~\cite{Antonio13das}. 

\begin{figure}
\centering
\includegraphics[width=9cm]{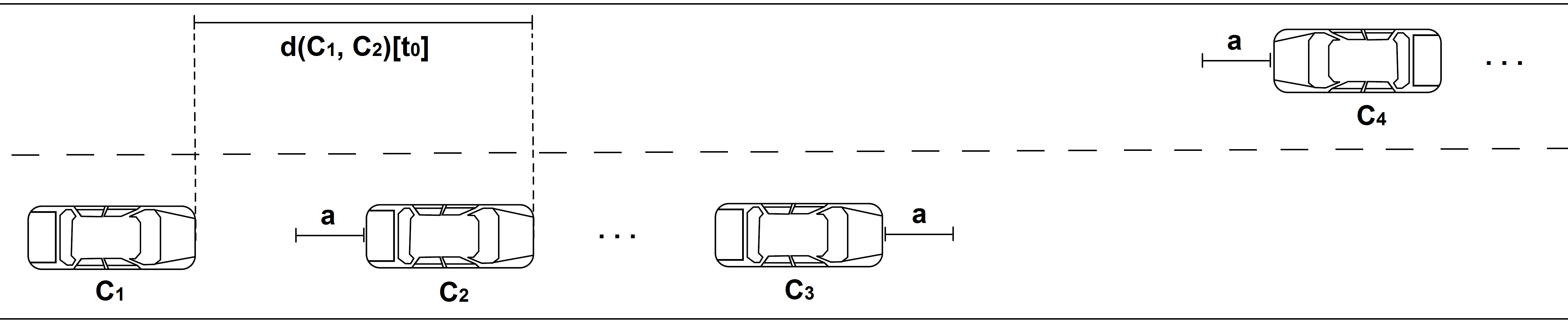}
\caption{$2^+$ overtaking situation.}
\label{fig:twoplus2}
\end{figure}

The situation in which one or more vehicles are in front on the overtaking vehicle, on the same lane, and an arbitrary number of vehicles can be present on the opposite lane is considered here.
The model assumes that the overtaking vehicle may start the maneuver at an arbitrary distance from the car in front. 
In Fig.~\ref{fig:twoplus2}, the car $c_1$ is the overtaking vehicle, as it is signaling its intension by signaling left. 
The three dots between $c_2$ and $c_3$ signify that an arbitrary number of vehicles may exist there. 
The left signal message is beaconed to the other agents in the vanet.

We consider the physical lengths of the cars to be important as vehicle lengths can vary significantly, depending on their type. 
Average personal automobiles usually have a length from 2 to 4 meters but the maximum length of a truck can exceed 25 meters. 
We denote them with $h_i$ for vehicle $c_i$. 
This information is obtained through Car-2-Car communication.


To calculate the safety level of an overtake, velocity and position of four vehicles are required:  the overtaking vehicle $c_1$, 
the closest vehicle $c_2$ and farthest vehicle $c_3$ from the same lane as the overtaker and the closest vehicle $c_4$  from the opposite lane.

Let $|c_i|$ be the velocity of the vehicle $c_i$, while $|c_i|_x$ and $|c_i|_y$ be the velocity components on $x$ and $y$-axis, respectively. 
We denote with $d_{ij}[\tau_k]$ the distance between vehicles $c_i$ and $c_j$ at time $\tau_k$. 
Here, $\tau_k$ represents the time interval since the overtaking maneuver started. 
We assume the distance between $c_i$ and $c_j$ is estimated at time $t_0$ when the overtaking maneuver starts based on vanets communication.
This distance can be calculated using equation~(\ref{eq:dist2}).

\begin{equation}
\label{eq:dist2}
d_{ij}[\tau_k] = d_{ij}[\tau_0] + (\tau_k \cdot ||c_i| - |c_j||)
\end{equation}


There are two steps for assessing the safety of the maneuver. 
Firstly, we need to calculate the required time to overtake $tto$. 
Secondly, the time to collision $ttc$ should be estimated.
$tto$ consists of the time required for lane-change and the time spent on the opposite lane. 
When starting to change the lane, vehicle $c_1$ will make an angle $\theta$ with the road lane. 
Vector decomposition~\cite{Antonio13das} is used to calculate the velocity components on $x$ and $y$-axis of vehicle $c_1$, 
using equations~(\ref{eq:vcc}) and~(\ref{eq:vcc2}).

\begin{align}
|{c_1}|_x &= |c_1| \cdot cos(\theta) \label{eq:vcc}\\
|{c_1}|_y & = |c_1| \cdot sin(\theta) \label{eq:vcc2}
\end{align}


The vehicle $c_1$ needs to travel a distance $L$ until the center of the opposite lane, where $L$ is equal to the width of a road lane. 
The time necessary $t_L$ to travel this distance is, given by equation~(\ref{eq:time1}). 
\begin{equation}
\label{eq:time1}
t_L = \frac{L}{|c_1|_y}
\end{equation}

The distance $d_x$ traveled in this time on $x$ axis is given by equation~(\ref{eq:dist1}).
\begin{equation}
\label{eq:dist1}
d_x = |c_1|+x \cdot t_L
\end{equation}

\begin{figure}
\centering
\includegraphics[width=9cm]{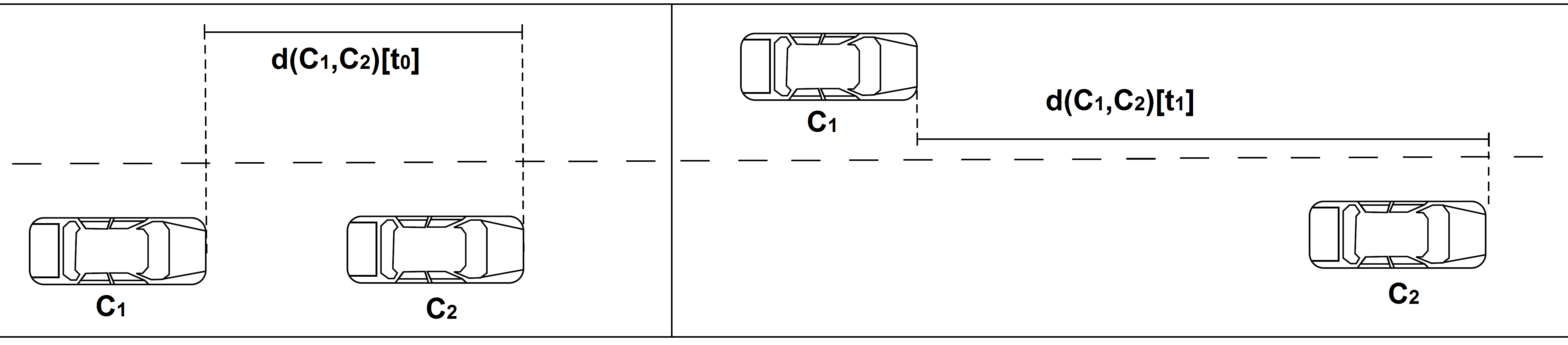}
\caption{Difference of distances between the vehicles $c_1$ and $c_2$ induced by the lane change.}
\label{fig:my2}
\end{figure}

The difference at time $\tau_0$ and at time $\tau_1$ is of distance between cars $c_1$ and $c_2$ given by equation~(\ref{cases}). 
If this difference is positive, the distance between the two vehicles has increased after the lane change (see Fig.~\ref{fig:my2}).

\begin{equation}
\label{cases}
\delta = |d_{ij}[\tau_1] - d_{ij}[\tau_0]| 
\end{equation}

\begin{figure}
\centering
\includegraphics[width=9cm]{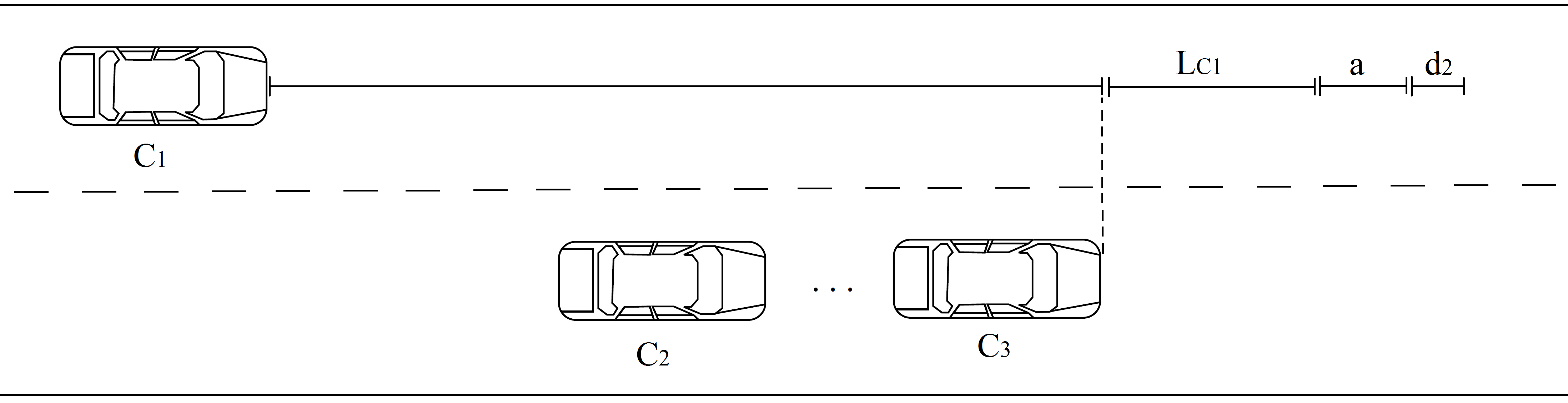}
\caption{Total distance traveled by $C_1$ on opposite lane.}
\label{fig:ttofinal}
\end{figure}

The distance to the farthest vehicle from the same lane, (that is vehicle $c_3$) is $d_{13}[\tau_1]$. 
Hence, the total distance traveled on the opposite lane (see Fig.~\ref{fig:ttofinal}) is given by: 

\begin{equation}
\label{eq:top}
\Delta(c_1) = d_{13} + h_1 + a + \delta
\end{equation}

Adding $\delta$ to the equation has the following effect: when $c_1$ changed lanes, the car's distance from $c_2$ changed, either increasing or decreasing. When returning to its lane, its distance to $c_3$ will change again, 
and $c_1$ will end up at exactly $a$ meters in front of $c_3$. 
We assume that velocities of $c_2$ and $c_3$ are equal.

The time required to cover this distance, relative to the vehicle $c_3$ is given by equation~(\ref{eq:t2}).

\begin{equation}
\label{eq:t2}
t(\Delta(c_1)) = \frac{\Delta(c_1)}{|c_1| - |c_3|}
\end{equation}

For the total time for the overtaking maneuver can be calculated with~\ref{eq:tto}. 
We add $t_x$ two times, because the lane change times are equal. 

\begin{equation}
\label{eq:tto}
TTO = t_L + 2 \cdot t_x
\end{equation}

The time to collision $TTC$ is: 

\begin{equation}
\label{eq:ttc}
TTC = \frac{d_{14}[\tau_0] }{|c_1| + |c_4|} 
\end{equation}

Based on \textit{TTO} and \textit{TTC}, safeness of the overtaking maneuver can be determined: if $TTO < TTC$ the overtaking is safe.

\subsection{Reducing the localization error}
For determining position of vehicle we can use a information from on-board sensors such as speed, steering angle. 
Most of modern car have on-board accelerometer, gyroscope and magnetometer (compass) that permit tracking the vehicle with high precision using dead reckoning techniques. Advantages of INS include high sample that is necessary for tracking high speed vehicle or during overtaking maneuvers,  autonomy - it work well in tunnel (when GPS data isn't accessible). Disadvantages of INS is increasing the position error with time.

Instead of INS the GPS is relatively slow $\sim10 SPS$
. GPS data have an error of $\pm{5}..\pm{10}m$ with Gaussian distribution but is stable in long time perspective. 
Using Inertial navigation in couple with GPS can improve significant the positioning data. 

Also for reducing error in positioning a Kalman filter can be applied on measured coordinate. Kalman filter is an effective procedure for combining data from noisy sensors to determine state of a system with uncertain dynamics \cite{gpsins}.
The filtration algorithm can be computed from next parts: 

1) predict the next system state $\hat{X}_{k}$  
\begin{equation}
\label{eq:EstVal}
{S}_{k}=AS_{k-1}+BU_{k}
\end{equation}

where $S_{k}$- state of vehicle at time k,
$A$- is the model (equation of motion) that predict new state,
$S_{k-1}$ is the state of vehicle at previous time k-1,
$B$ is the model that predict what changes based on commands to the vehicle (increase in throttle or steering);
$U_{k}$ is the command input at time k.

2) Project the error covariance ahead:
\begin{equation}
\label{Pk}
P_{k} = AP_{k-1}A^{T}+Q
\end{equation}
where $P_{k-1}$ previous error value, $Q$ -covariance of error noise - describe the distribution of noise.

3) Computing the Kalman Gain:
\begin{equation}
\label{Kgain}
K_{k} = \dfrac{P_{k-1}H^{T}}{HP_{k-1}H^{T}+R}
\end{equation}
where $K$ Kalman gain, $H$ - the model of how sensor reading reflect the vehicle state (function to go from sensor reading to state vector) $R$- describe the noise in sensor reading.

4) Update the estimate with measurements from sensor:
\begin{equation}
\label{XK}
S_{k}=S_{k-1}+K(Z_{k}-H{S}_{k-1})
\end{equation}
where $S_{k}$ is the state at time $k$ and output of filter, $S_{k-1}$ is the estimate of state we did previously, $Z_{k}$

If our case a current state is represented by vector $S$:
\[ S_{k}= \left( \begin{array}{c}
pos_{x} \\
pos_{y} \\
vel_{x}	\\
vel_{y}\end{array}\right) \] 
$pos_{x}$, $pos_{y}$- coordinates and $vel_{x}$, $vel_{y}$ is the velocity of vehicle.

\[ A= \left( \begin{array}{cccc}
1 & 0 & \Delta t & 0 \\
0 & 1 & 0 & \Delta t  \\
0 & 0 & 1 & 0 \\
0 & 0 & 0 & 1 \\\end{array}\right) \] 
  
\[ B= \left( \begin{array}{cc}
\dfrac{1}{2}\Delta t & 0     \\
0 & \dfrac{1}{2}\Delta t 	\\
\Delta t & 0 				\\
0 & \Delta t \end{array}\right) \] $\Delta t$, - time interval in update vehicle state.
   
\[ U_{k}= \left( \begin{array}{c}
a_{x}     \\
a_{y} 	\\
\end{array}\right) \] $a_{x}$, $a_{x}$ - projection??? of vehicle acceleration on $X$ and $Y$ axis.
\textbf{}

Based on current state we can predict the next state:
\[ \hat{S}_{k}= \left[\begin{array}{c}
pos_{x} \\
pos_{y} \\
vel_{x}	\\
vel_{y}\end{array}\right]
\left[\begin{array}{cccc}
1 & 0 & \Delta t & 0 \\
0 & 1 & 0 & \Delta t  \\
0 & 0 & 1 & 0 \\
0 & 0 & 0 & 1 \\\end{array} \right]+
\left[ \begin{array}{cc}
\dfrac{1}{2}\Delta t & 0     \\
0 & \dfrac{1}{2}\Delta t 	\\
\Delta t & 0 				\\
0 & \Delta t \end{array}\right]
\left[ \begin{array}{c}
a_{x}     \\
a_{y} 	\\
\end{array}\right] \] 
 \[=
\left[\begin{array}{c}
pos_{x}+vel_{x}\Delta t+a_{x}\dfrac{1}{2}\Delta t^{2} \\
pos_{y}+vel_{y}\Delta t+a_{y}\dfrac{1}{2}\Delta t^{2}  \\
vel_{x}+a_{x}\dfrac{1}{2}\Delta t
vel_{y}+a_{y}\dfrac{1}{2}\Delta t
\end{array}
\right]
\]
 
The initial state covariance matrix can be written as:
\[ P_{k0}= \left( \begin{array}{cccc}
\sigma^{2}_{pos_{x}}  &  \sigma_{pos_{x},pos_{y}}  &  \sigma_{pos_{x},vel_{x}}  &  \sigma_{pos_{x},vel_{y}} \\
\sigma_{pos_{x},pos_{y}} & \sigma^{2}_{pos_{y}}  & \sigma_{pos_{y},vel_{x}} & \sigma_{pos_{y},vel_{y}}  \\
\sigma_{pos_{y},vel_{x}} & \sigma_{pos_{y},vel_{x}} & \sigma^{2}_{vel_{x}}  & \sigma_{vel_{x},vel_{y}} \\
\sigma_{pos_{x},vel_{y}} & \sigma_{pos_{y},vel_{y}} & \sigma_{vel_{x},vel_{y}} & \sigma^{2}_{vel_{y}}  \\\end{array}\right) \]

The next state covariance matrix is calculated using relation (\ref{Pk})

Using the estimated covariance matrix $P_{k}$ and the measured coordinates from GPS receiver $\check{S}= \left[ 
\begin{array}{c}
\check{pos_{x}}	\\
\check{pos_{y}}	\\
\check{vel_{x}}	\\
\check{vel_{y}}	\\
\end{array}
\right]$
We can calculate Kalman gain $K_{k}$ by (\ref{Kgain}) and in final approximate the new state using relation (\ref{XK}).

Each vehicle $v_i$ performs inter-vehicle distance measurements $d^s_{ij}$, and takes a reading of its own velocity $s^s_i$. 
This information is then shared with all vehicles in $\mathcal{V}_o$.

\section{Employing safety overtaking by multi-agent systems}
\label{sec:agents}

This section details the communication protocol and the experiments run based on multi-agent systems.

 \subsection{Overtaking communication protocol}

This section presents the communication protocol that precedes the overtaking maneuver. 
The protocol aims twofold: 
Firstly, to obtain a confirmation from the lead vehicle that there is not another overtaking car from the opposite lane. 
Secondly, to notify other traffic participants of the overtaking that is about to take place. 
The first goal is a workaround for the limited communication range problem in vehicular networks. 
The intended effect of the second goal is that vehicles will keep a constant speed during the maneuver and they will not try to overtake at the same time. 

\begin{figure}
\centering
\includegraphics[width=4.5cm]{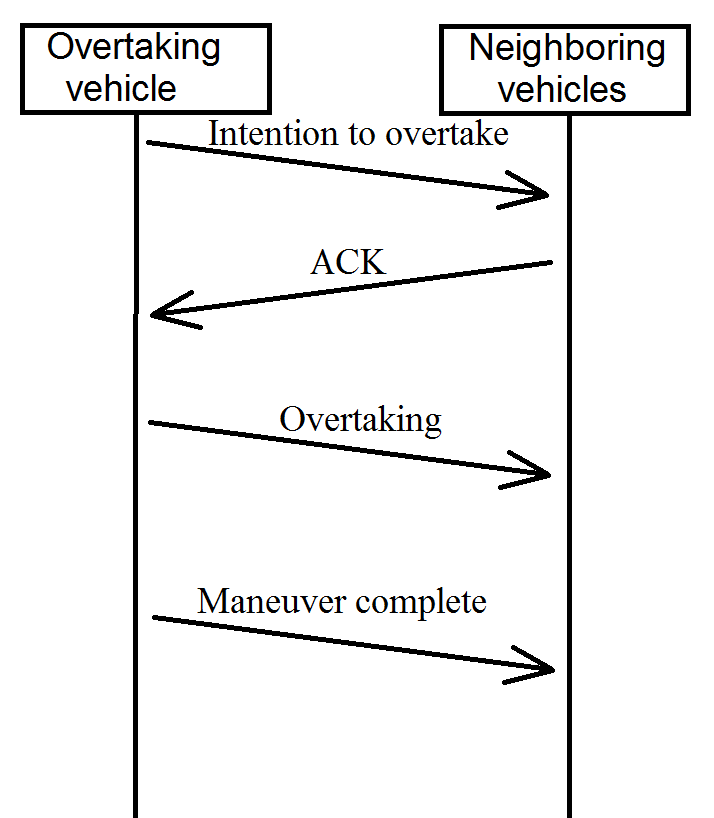}
\caption{Proposed protocol for the overtaking maneuver.}
\label{fig:protocolOVT}
\end{figure}

If the safety module predicts a safe overtaking possibility, an overtaking intention protocol is initiated (see Fig.~\ref{fig:protocolOVT}).
An \textit{acknowledgment/negative acknowledgment} (\textit{ACK-NACK}) protocol is enacted in which the destination is required to respond back, once a message is received from the sender.
Normally, an acknowledgment-based (ACK) protocol is desired, because it minimizes the bandwidth usage of the network and lowers the communication overhead. However, since package loss may be a problem in vanet-based safety applications, we need to distinguish if a packet is lost (no response from a node) or a negative response is returned, indicating a dangerous situation.

\begin{figure}
\centering
\includegraphics[width=8.5cm]{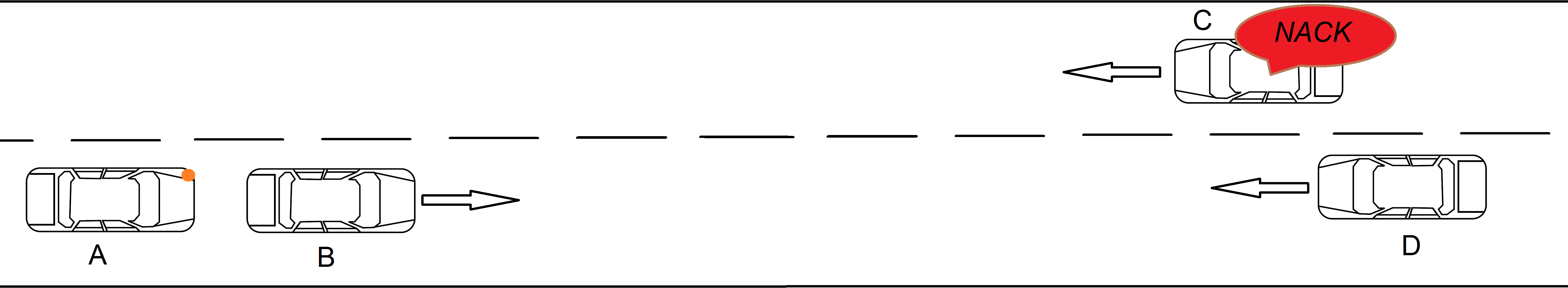}
\caption{A NACK response signals that an overtaking maneuver is already in progress. The protocol warns vehicle $A$ to not engage in its intended overtaking maneuver.}
\label{fig:nack_msg}
\end{figure}

The protocol in Fig.~\ref{fig:protocolOVT} is initiated by the overtaking car that sends an event-based message, to notify other traffic participants of its intention. 
Vehicles receiving this message should reply with an acknowledge (ACK) or negative-acknowledge (NACK).
A NACK packet is returned by a traffic participant if that node has knowledge of some dangerous situation. 
An example is presented in Fig.~\ref{fig:nack_msg} where an overtaking maneuver is already in progress. 
This is useful when the overtaking vehicle $A$ is not in the communication range with the opposite overtaking vehicle $D$. 
Otherwise, an ACK message is sent in response to the $A$' intention. 
By sending an ACK, vehicles notify the overtaking car that they will maintain a constant speed throughout the maneuver.

\begin{figure}
\centering
\includegraphics[width=5.5cm]{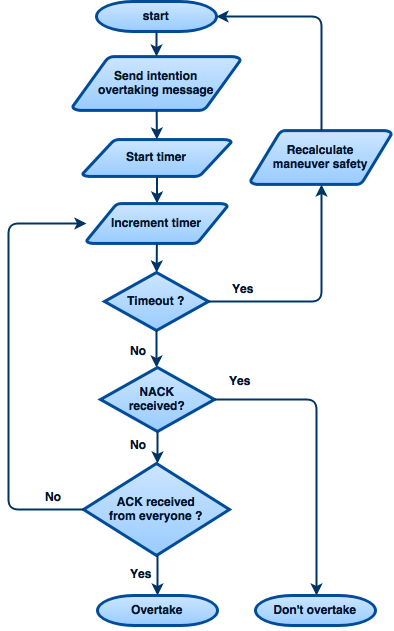}
\caption{Logic flow of the overtaking protocol.}
\label{fig:flow1}
\end{figure}

After sending the intention message, the overtaking agent has to monitor the time it takes for all neighbors to respond. This needs to be done since after a period of time the initial overtaking calculations will no longer be valid. 
If a timeout period has been reached, the agent needs to perform the calculations again.
A general diagram of the protocol phases is presented in Fig.~\ref{fig:flow1}.

\begin{figure}[ht]
\centering
\includegraphics[width=8cm]{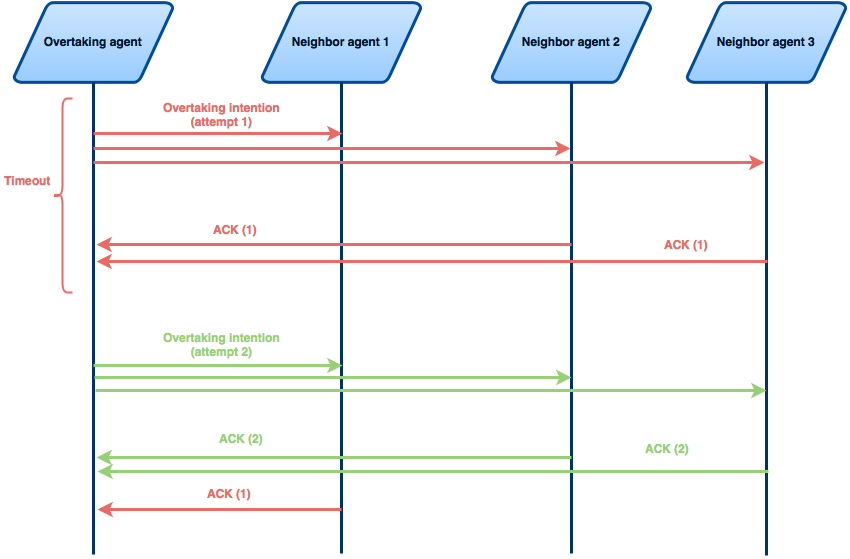}
\caption{Possible inconsistencies introduced due to delays or lost messages.}
\label{fig:proto3}
\end{figure}

Introducing the concept of a timeout/resend technique poses the following problem: 
Suppose the overtaker agent has issued an overtaking request message (see Fig.~\ref{fig:proto3}). 
It receives acknowledgments from all neighbors, except one, which is delayed. 
The agent then resends the message and as the new \textit{ACK}s are received, the acknowledgment from the previous request is also processed. 
The overtaking agent may mistake this for a more recent response to the request. 
This result is undesirable because it may induce an error in the system. 
The agent may decide that a situation is safe, when in fact it is not, or the other way around. 

A solution to this problem is to introduce request numbering. 
First, a message with a request identifier 0 is sent. 
If the request time out, a new message with identifier 1 is sent, and so on. 
When responding to such requests, agents will also place the request number in the message, so that the overtaking agent can determine to which request the response is sent. 

\subsection{Multi-agent vehicular simulation}

\begin{figure}
\centering
\includegraphics[width=8.5cm]{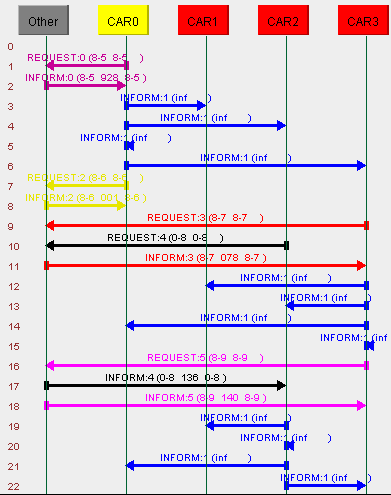}
\caption{The Sniffer agent monitoring message exchanges in JADE.}
\label{fig:sniffer}
\end{figure}

To model automobiles in our VANET as intelligent agents, Jade is used.  
We had to map the DSRC messages in VANETs to the FIPA-ACL messages in JADE. 
FIPA-ACL communicative acts were used  to simulate actual communication exchange in VANETs as illustrated by the Sniffer agent in Fig.~\ref{fig:sniffer}.



\begin{figure}[ht]
\centering
\includegraphics[width=8.5cm]{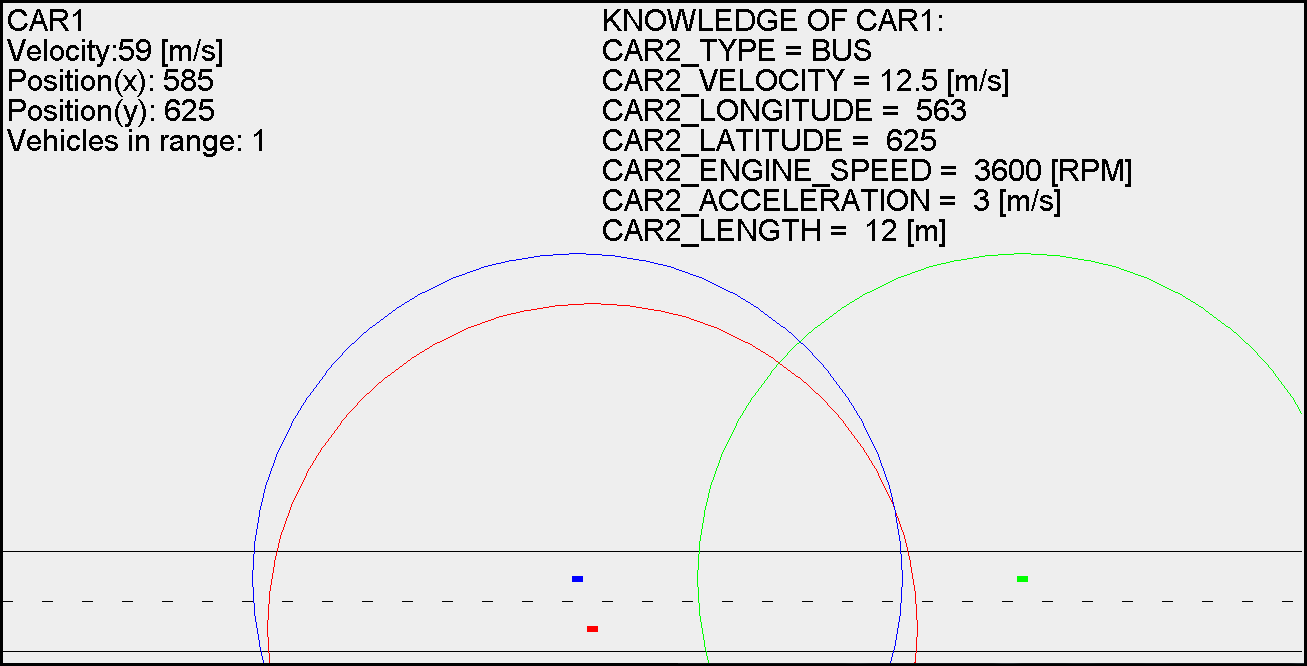}
\caption{Graphic interface for visualization of the overtaking scenario.}
\label{fig:gui}
\end{figure}

We define a \textit{mis-prediction} of the system as an inaccurate output (decision) taken after the algorithm finishes its execution (i.e. a safe prediction was made, when the maneuver was not in fact safe, or the other way around).
The experiments were centered around the concept of communication distance and how it impacts the mis-predicted number of maneuvers. 
Tests were performed on a scenario with four vehicles (recall Fig.~\ref{fig:twoplus2}). 
Velocities and inter-vehicle distances were modified randomly. 
Communication range was varied from 100 to 1000 meters with 50 meter increments. 
For each increment, 500 maneuvers were simulated with random value intervals each time. 
Thus, the mis-prediction count represented in the following graphs on the $y$-axis is from a total of 500 cases.


\paragraph{Velocity variation}

\begin{figure}
\centering
\includegraphics[width=8cm]{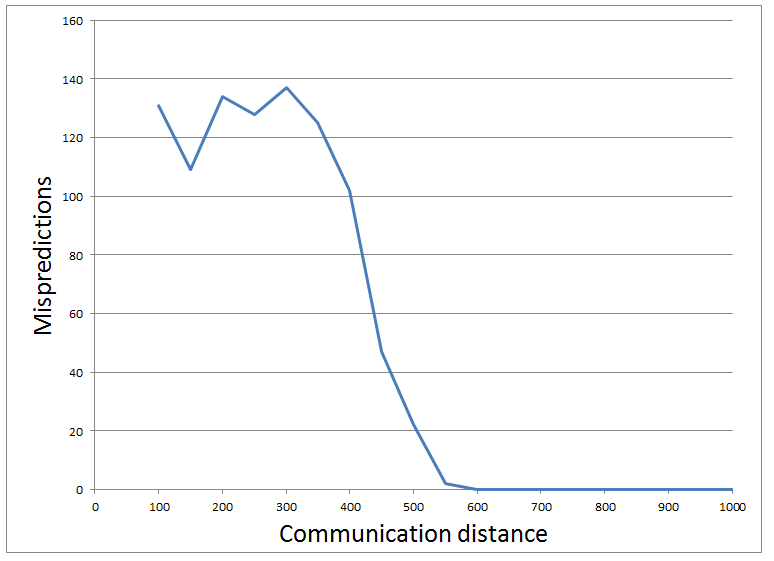}
\caption{Low velocities: $|c_0|,|c_3|\in[50..60]$, $|c_1|,|c_2|\in[40..50]$.}
\label{fig:low_speeds}
\end{figure}

The focus here is on how the vehicle velocities impact the number of mis-predictions. 
The results in Fig.~\ref{fig:low_speeds} shows mis-prediction levels for relatively low vehicle velocities.

For low velocities (Fig.~\ref{fig:low_speeds}), we used random values in interval 50 to 60 km/h for the cars $c_0$ and $c_3$ and 40 to 50 km/h for the vehicles $c_1$ and $c_2$.

\begin{figure}
\centering
\includegraphics[width=8cm]{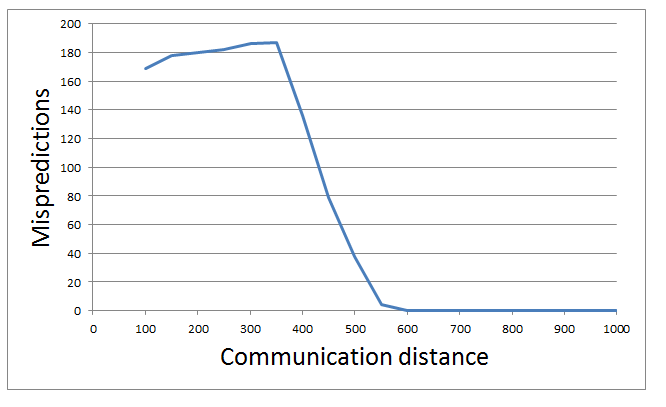}
\caption{Medium velocities: $|c_0|,|c_3|\in[70..80]$, $|c_1|,|c_2|\in[60..70]$.}
\label{fig:medium_speeds}
\end{figure}

For medium velocities (Fig.~\ref{fig:medium_speeds}), we used random values in interval 70 to 80 km/h for cars $c_0$ and $c_3$ and 60 to 70 km/h for cars $c_1$ and $c_2$.

\begin{figure}
\centering
\includegraphics[width=8cm]{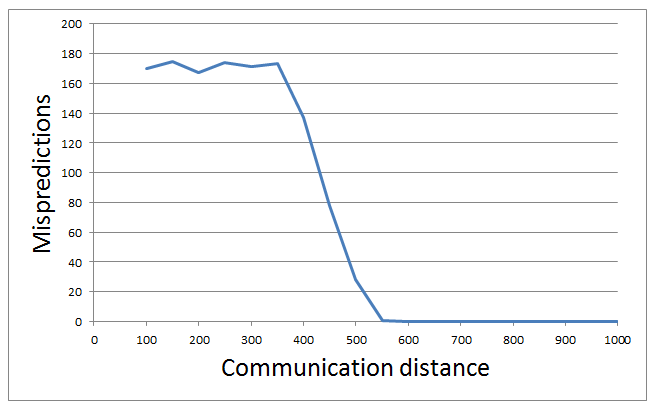}
\caption{High velocities: $|c_0|,|c_3|\in[70..80]$, $|c_1|,|c_2|\in[60..70]$.}
\label{fig:high_speeds}
\end{figure}

For high velocities (Fig.~\ref{fig:high_speeds}) we used random values in interval 100 to 120 km/h for cars $c_0$ and $c_3$ and 80 to 90 km/h for cars $c_1$ and $c_2$.

\paragraph{Distance variation}
Focus here is on how distance between cars affect the mis-prediction ratio. 
The vehicle $c_3$'s position was always chosen in the interval $[cd-15m,cd+15m]$, where $cd$ is the communication distance.

\begin{figure}
\centering
\includegraphics[width=8cm]{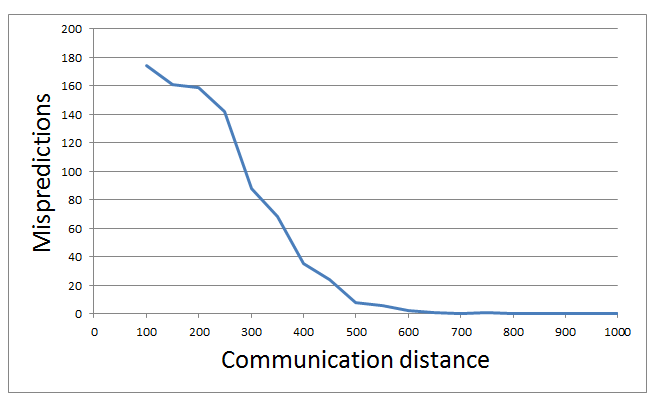}
\caption{Short distances: $d_{01} = [1..8][m]$, $d_{02}=[9..17][m]$.}
\label{fig:small_distances}
\end{figure}

For short distances (Fig.~\ref{fig:small_distances}), values were generated randomly in the following intervals: $d_{01} = [1..8][m]$ and $d_{02}=[9..17][m]$.

\begin{figure}
\centering
\includegraphics[width=8cm]{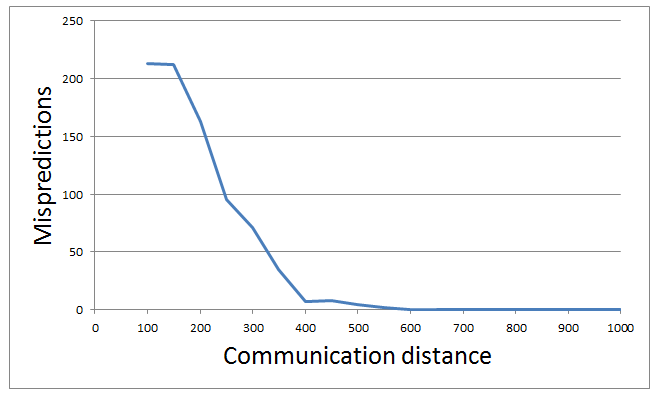}
\caption{Medium distances: $d_{01} = [5..14][m]$ and $d_{02}=[15..24][m]$.}
\label{fig:medium_distances}
\end{figure}

For medium distances (Fig.~\ref{fig:medium_distances}), values were generated in the intervals: $d_{01} = [5..14][m]$ and $d_{02}=[15..24][m]$..

\begin{figure}
\centering
\includegraphics[width=8cm]{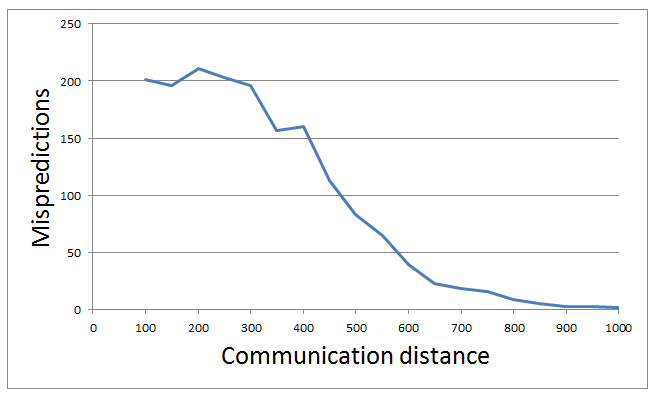}
\caption{Long distances: $d_{01} = [10..34][m]$, $d_{02}=[35..44][m]$.}
\label{fig:high_distances}
\end{figure}

For long distances (Fig.~\ref{fig:high_distances}), values were generated in the intervals: $d(C_0,C_1) = [10,34][m]$, $d(C_0,C_2)=[35,44][m]$.

The results in all experiments show a general trend of mis-predictions to drop significantly between 400 and 600 meters of communication range.

\section{Discussion and Related Work}
\label{sec:discussion}
Vehicular agents are empowered with domain knowledge and they can perform  geospatial  and  temporal  reasoning in ~\cite{groza2014multi}. 
An ontology for the vehicular network domain has been developed for supporting agents reasoning in vanet-based cooperative applications~\cite{groza2014multi}.  
A different line of research~\cite{hsuinternet} is based on mobile agents and norm-aware agents. 
In this paper, we assume communication through vanet technologies, aiming to be consistent with the IEEE802.11p standard. 
A trust model in~\cite{minhas2010intelligent} employs a penalty in case of misleading reports. 
Experiments with altruist/selfish agents in vanets~\cite{lo2012collaborative} 
shown that average speed increased when the
vehicular agents cooperate. 
A vanet-based emergency vehicle warning system has been developed in~\cite{groza2014towards}. 
In line with~\cite{ilarri2015semantic} which argue on semantic exchange of
mobility data and knowledge among moving objects, we will take a step towards formalizing
vehicular-related knowledge and semantic interchange of events.

Minimizing localization error through Car-2-Car communication~\cite{toledo2009cooperative} or based on the Extended Kalman Filter~\cite{efatmaneshnik2010cooperative} is an important aspect in the context of safety overtaking. 
The problem of cooperative localization has been approached from various perspectives such as  Extended Kalman Filter~\cite{efatmaneshnik2010cooperative, de2013driver} or ad-hoc trilateration~\cite{parker2007cooperative}. 
The system uses signal strength based inter-vehicle distance measurements, road maps, vehicle
kinematics, and Extended Kalman Filtering to estimate relative positions of vehicles in a cluster.  
The road map is used to ensure that position estimates are within the road boundaries.
Distance-bounding protocols are used to determine the upper bound on the physical distance to another vehicle, with the disadvantage that these protocols rely on roadside infrastructure. 
In our case the cluster is represented by the vehicles implied in the overtaking event. 

The standard deviation of GPS-based positioning error is $\sigma_P=7m$ and 
standard deviation of ranging measurements error between two vehicles $\sigma_R=5m$ in~\cite{yao2011improving} . 
The ranging measurements are periodically exchange location, speed, and other kinematic information.
Extensive on-the road experiment was performed~\cite{amoroso2012gps} to test GPS errors for an accident warning system in VANETS, using a Sirf Star III GPS receiver, with a accuracy of 5 to 10 m, representing the accuracy of commercial GPS receiver. 
In the measurements performed in the real life scenario the absolute values of the average $ep$ 
($ep$ representing the actual error on the position of a vehicle) were in the order of magnitude of a medium vehicle (about 4m).
The accuracy of on-board speed and orientation sensors (gyroscope) are considered in~\cite{lee2012rfid, lee2009rf}, having an error range of around $\pm3\sim\pm10\%$, respectively 10?/sec of orientation error at maximum. That is $\sigma_S = \pm3\sim\pm10\%$ and $\sigma_G = 10?/sec$. 
The effect of speed error on positioning accuracy (PA) using Kalman filter based CP techniques is described in~\cite{efatmaneshnik2011cooperative} 
$\sigma_S = 0.1m/s$ results in a $PA < 1m$.
Other studies~\cite{shladover2006analysis} ignore the speed error component, considering the wheel speed information to be adequate accurate.
For overtaking situations, Shladover et al. in~\cite{shladover2006analysis} consider three criteria for a successful collision warning system:
(1) a minimum threshold lateral distance of 4.8m between vehicles centers,
(2) longitudinal positions should be in the range of 1.5m to 6m, and 
(3) accurately identify vehicles speeds.
Using network improvements methods to reduce communication overhead in VANETS - due to periodically exchange of messages (for example ranging measurements), can reduce packet loss and improve positioning accuracy. 
In~\cite{yao2011improving} several methods have been studied: piggyback, compression, reducing broadcasting interval and network coding. 
The combination of network improvements methods with the CP technique brought a 40\% improvement in positioning accuracy.

\section{Conclusion}
\label{sec:conclusion}
We developed a warning system for assisting drivers during overtaking maneuvers. 
We exploited multi-agent systems and Car-2-Car communication in vehicular networks.

The proposed framework can be used to further develop various communication protocols for vehicular-based ACL messages towards plausible reasoning~\cite{groza2012plausible} on vehicular data streams 

%
\IEEEpeerreviewmaketitle


\bibliographystyle{IEEEtran}      
\bibliography{nlq,thesis}

\end{document}